\pdfoutput=1

\documentclass[11pt]{article}

\usepackage[final]{nlp4MusA}

\usepackage{times}
\usepackage{latexsym}
\usepackage{booktabs}

\usepackage[T1]{fontenc}

\usepackage[utf8]{inputenc}

\usepackage{microtype}

\usepackage{inconsolata}

\usepackage{graphicx}

\usepackage{subcaption}

\title{Evaluation of pretrained language models on music understanding}

\author{Yannis Vasilakis \\
  Queen Mary University of London \\
  \texttt{i.vasilakis@qmul.ac.uk} \\\And
  Rachel Bittner \\
  Spotify \\
  \texttt{rachelbittner@spotify.com} \\\And
  Johan Pauwels \\
  Queen Mary University of London \\
  \texttt{j.pauwels@qmul.ac.uk}}

\begin{document}
\maketitle
\begin{abstract}
Music-text multimodal systems have enabled new approaches to Music Information Research (MIR) applications such as audio-to-text and text-to-audio retrieval, text-based song generation, and music captioning. Despite the reported success, little effort has been put into evaluating the musical knowledge of Large Language Models (LLM). In this paper, we demonstrate that LLMs suffer from 1) prompt sensitivity, 2) inability to model negation (e.g. ``rock song without guitar''), and 3) sensitivity towards the presence of specific words. We quantified these properties as a triplet-based accuracy, evaluating the ability to model the relative similarity of labels in a hierarchical ontology. We leveraged the Audioset ontology to generate triplets consisting of an anchor, a positive (relevant) label, and a negative (less relevant) label for the genre and instruments sub-tree. We evaluated the triplet-based musical knowledge for six general-purpose Transformer-based models. The triplets obtained through this methodology required filtering, as some were difficult to judge and therefore relatively uninformative for evaluation purposes. Despite the relatively high accuracy reported, inconsistencies are evident in all six models, suggesting that off-the-shelf LLMs need adaptation to music before use.
\end{abstract}

\section{Introduction}

The capability of Large Language Models (LLM) to obtain informative context-dependent word embeddings with long-range inter-token dependencies showed that they can be used effectively to encode knowledge from several domains without manually curating datasets.

During the last 5 years, the scientific community combined audio-based Deep Neural Networks (DNN) with LLMs to form audio-text models, leading to improved performance on several music applications such as audio-to-text retrieval and text-to-audio retrieval~\cite{mulan_a_joint_embedding_qinqing_2022, muscall_manco_2022, laion_clap_Wu_2023}, music captioning~\cite{llark_gardner_2024, muscaps_manco_2021} and text-based song generation~\cite{museformer_botao_2022}.

LLMs are usually used pretrained and off-the-shelf~\cite{muscall_manco_2022, mulan_a_joint_embedding_qinqing_2022}. While datasets for semantic similarity of general language~\cite{semeval-2024-international} are available, we are not aware of any such datasets for music. Therefore, LLMs haven't been thoroughly evaluated on their musical knowledge and potential issues might be obscured.

In this paper, we quantify musical knowledge in LLMs using triplets obtained through an ontology and report three shortcomings when used off-the-shelf. We leverage Audioset, a hierarchical ontology, to extract the triplets of (anchor, positive, negative) format. The anchor label is chosen arbitrarily from the ontology, a similar label is selected as the positive, and a relatively less similar label as the negative term of the triplet. We quantify the relative similarity using the ontology-based distance between pairs of labels. Thus, we evaluate LLM's musical knowledge by comparing the relative similarity between anchor-positive and anchor-negative labels. We collected 13633 Music Genre and 37640 Music Instrument triplets. We evaluated the sensitivity of LLMs to 20 different musically informed prompts and their inability to model negation. Finally, we report performance improvements when both labels and their definitions are used.

Both code snippets and sets of triplets used are made publicly available for reproducibility reasons~\footnote{https://github.com/YannisBilly/Evaluation-of-pretrained-language-models-on-music-understanding}.

\section{Related Work}

\subsection{BERT}

Bidirectional Encoder Representations from Transformers (BERT)~\cite{devlin-etal-2019-bert, 10.1162/tacl_a_00477} is the backbone for many Natural Language Processing (NLP) applications such as translation~\cite{Xu2021BERTMO}, text summarization~\cite{text_summarization_encoders_liu_2019}, and others. These systems were trained with unstructured large corpora through masked word and next-sentence prediction, without the need for curated datasets.

BERT provides a context-dependent token-based embedding vector but doesn't calculate independent sentence embeddings. This means that sentence embeddings need to be calculated as a function of the token embeddings at inference time. Obtaining the latter is not straightforward~\cite{evaluation_of_bert_sentence_Choi_2021, factors_affecting_sentence_similarity_Alian_2020} and several different approaches have been proposed. The most frequent, better-performing method is averaging the token embeddings in different layer depths. Another one is using the [CLS] token, obtaining subpar performance~\cite{on_sentence_embeddings_Li_2020}. We focus on the first approach as the most prominent but highlight that calculating sentence embeddings is still an active research topic~\cite{XU2024102065, app13063911}.

\subsection{Large Language Models in Music Information Research}

Transformer-based models have been introduced in several applications. Zero-shot classification utilizes word embeddings to infer a classifier on unseen classes based on the similarity of the new class label with the labels of the known classes~\cite{Du2023JointMA}. Audio-to-text and text-to-audio retrieval is successful in aligning audio and text embeddings using music/caption pairs~\cite{muscall_manco_2022, mulan_a_joint_embedding_qinqing_2022}. Automatic music caption uses music embeddings to condition an LLM~\cite{muscaps_manco_2021, llark_gardner_2024} to generate music descriptions. Lastly, sentence similarity has been used to weigh intra-caption similarity in contrastive loss functions~\cite{muscall_manco_2022, oord2018representation}.



\section{Evaluation of language models on musical knowledge}

As far as we are concerned, a linguistic evaluation dataset of musical knowledge doesn't exist apart from language-based artist similarity~\cite{natural_language_knowledge_discovery_Oramas_2018, a_semantic_based_artist_similarity_Oramas_2015}.

Information used for semantic similarity is usually scraped from websites and we argue that this information is not directly useable. Generally, these websites highlight the history of the queried label without juxtaposing related concepts, audio attributes or providing slang labels and abbreviations. Also, their massive size can hinder inspection and therefore, reduce their value as evaluation sets.

We argue that an evaluation dataset needs to be cleaned and inspected thoroughly before increasing its size. This hasn't been done in captioning and tagging datasets, as most are weakly annotated and have highly noisy annotations~\cite{the_effects_of_noisy_labels_Choi_2018}.

Therefore, we chose to utilize an ontology with less than 200 musical labels which have a manageable size, can be manually inspected and filtered. However, we need to acknowledge that most existing ontologies are far from being exhaustive. We drew inspiration from the Semantic Textual Similarity task~\cite{semeval-2024-international,dong2021parasci, paraphrasing_dataset_Wahle_2022} that contains pairs of sentences and their degree of similarity but proposed a method of obtaining such sentences automatically leveraging a taxonomy.

We evaluated 6 general-purpose Transformer-based models~\cite{Reimers2019SentenceBERTSE} for sentence similarity using musical terminology. In detail, a global average pooling layer is appended on top of the final layer and the sentence embedding is calculated as the mean of the respective token embeddings. The models used are \textit{MPNet, DistilRoBERTa, MiniLM} and \textit{ALBERT} trained on different corpora. More information about the models is provided in appendix section~\ref{appendix:language_models_used} and tables~\ref{tab:prompt_sensitivity_results},~\ref{tab:using_definition}.

\subsection{Audioset and its ontology}
Large-scale annotated datasets have been essential for Computer Vision. Drawing inspiration from this, Audioset~\cite{7952261} was proposed which has $\approx 1.79$ million 10-second long audio snippets scraped from YouTube, annotated with a hierarchical ontology of 632 audio classes.

The creation of their taxonomy focused on two properties: (1) labels must be recognizable by typical listeners without additional information beyond the label, and (2) the taxonomy must be comprehensive enough to describe most real-world recordings adequately. After finalizing the taxonomy, annotators were given a 10-second audio clip and a label. They had to choose from ``present'', ``not present'', or ``unsure'' to indicate whether the audio and label used were positively, negatively, or uncertainly related, respectively.

In this paper, we use the Audioset sub-tree of Music~\footnote{Visualization: http://www.jordipons.me/apps/audioset/}. Due to the unitary depth of most child nodes (e.g. Music mood), we will only include the sub-trees of ``Musical Instrument'' and ``Music Genre''. A deficiency of using a tree is that inter-category relations cannot be modeled (e.g. ``Rock music'' and ``Guitar''). The triplet-based evaluation methodology can be extended to other graph structures and elaborate ontologies (e.g. WordNet~\cite{wordnet_miller_1995}), as well as include intra-category relations (e.g. ``Rock music', ``Electric guitar'', ``Viola'').

The ``Musical Instrument'' taxonomy has a maximum depth of 4, encompassing most instrument families, including classical, modern, and non-western instruments. Although it does not separate playing techniques from instruments (e.g., ``electric guitar'' and ``tapping''), omits some instruments (e.g., ``viola'' from ``bowed string instruments'') and contains vague concepts (``Musical ensemble''), the taxonomy remains well-defined and free of ambiguous labels.

The ``Musical genre'' taxonomy has a maximum depth of 3, covering Western music with detailed categorization of contemporary genres (e.g., ``Grime music''), as well as folk and non-Western genres. However, it lacks nuance in classical music, only including opera.

\subsection{Triplet-based musical knowledge quantification}
\label{subsection_triplet_based_musical_knowledge_quantification}
To curate the music knowledge corpus for LLM evaluation, we leverage the aforementioned sub-trees of the Audioset ontology and generate triplets. Specifically, we form triplets of an anchor, a positive and a negative label. The positive and negative labels are defined relative to their semantic similarity with respect to the anchor label.  If the anchor is more similar to label 1 than label 2, label 1 is the positive and label 2 is the negative label. This method can encode abstract relationships between labels, including comparisons between non-homogeneous labels (e.g., ``happy music'', ``rock music'', ``reggae music'') but is left for future work as it requires more elaborate ontologies.

We use the distance between the labels based on each tree to quantify their relative similarity. A valid triplet is defined as one where the anchor-positive is less than the anchor-negative distance. After obtaining the valid triplets, we manually inspect them and remove the ones that are ambiguous, vague or too difficult to judge\footnote{Removed triplet cases are provided in Appendix table~\ref{tab:ambiguous_triplets}}.

Finally, we are left with 13633 Genre triplets and 37640 Instrument triplets that will be evaluated separately. Despite the manual inspection, it is important to declare that the dataset is biased toward authors' knowledge of Western music and some triplets might have been erroneously left out.

\subsection{Experiments and results}


After obtaining the sentence embedding using triplets, cosine similarity will be used to evaluate the relative semantic similarity. Anchor-positive and anchor-negative cosine similarity will be compared and a triplet will be regarded as correct if the first is greater than the second. A thorough analysis of the results is provided in the appendix chapter~\ref{appendix:detailed_experiment_results}. Finally, the accuracy of correct triplets will be calculated and reported.

\subsubsection{Prompt sensitivity}
\label{subsubsection_prompt_sensitivity}
Wrapping queried labels in a prompt is useful~\cite{Radford2021LearningTV} but we are not aware of a thorough analysis of the performance variance concerning different prompts. As a result, we used 20 musically informed prompts. The exact wording of the prompts is provided in appendix~\ref{appendix:prompt_sensitivity}. Several words as ``music'', ``recording'' or ``sound'' have been used, to simulate human music captions/descriptions.

The standard deviation reported is relatively high for every case apart from the paraphrased-MiniLM model as presented in table~\ref{tab:prompt_sensitivity_results}. As the prompts do not provide additional information, it can be argued that the models are moderately sensitive to the prompts and ``musical'' words added can be useful. Lastly, the best model according to model size and performance is paraphrased-ALBERT.

\begin{table*}
  \centering
  \begin{tabular}{lcccc}
    \toprule
    \textbf{} & \multicolumn{2}{c}{\textbf{Prompts}} & \multicolumn{2}{c}{\textbf{Negation}}\\
    \textbf{Models} & \textbf{Instruments} & \textbf{Genres} & \textbf{Instruments} & \textbf{Genres}\\
    \midrule
    mpnet-base      & $71.3 \pm 3.7$ & $76.4 \pm 2.3$ & $41.1 \pm 3.7$ & $43.2 \pm 3.8$\\
    distilroberta   & $62.4 \pm 2.4$ & $69.6 \pm 2.6$ & $37.2 \pm 3.6$ & $42.3 \pm 3.4$\\
    MiniLM-L12-v2   & $62.7 \pm 2.3$ & $70.9 \pm 2.3$ & $33.8 \pm 6.5$ & $37.3 \pm 6.9$\\
    MiniLM-L6       & $65.8 \pm 2.7$ & $70.5 \pm 1.6$ & $37.4 \pm 5.8$ & $41.4 \pm 5.8$\\
    Para-albert     & $69.6 \pm 3.2$ & $66.5 \pm 1.7$ & $33.4 \pm 5.8$ & $35.6 \pm 5.7$\\
    Para-MiniLM-L3  & $63.2 \pm 2.7$ & $66.9 \pm 0.8$ & $29.0 \pm 6.7$ & $34.3 \pm 5.0$\\
    \bottomrule
  \end{tabular}
  \caption{Presenting the percentage of correctly inferred triplets for Instruments and Genres respectively. Prompt sensitivity showcased from high standard deviation along 20 prompts. Also, Transformer-based models cannot model negation as the accuracy obtained is worse than random.}
  \label{tab:prompt_sensitivity_results}
\end{table*}

\begin{table*}
  \centering
  \begin{tabular}{lcccc}
    \toprule
    \textbf{} & \multicolumn{2}{c}{\textbf{Instrument Definitions}} & \multicolumn{2}{c}{\textbf{Genre Definitions}}\\
    \textbf{Models} & \textbf{Definition + Label} & \textbf{Definition - Label} & \textbf{Definition - Label} & \textbf{Definition + Label}\\
    \midrule
    mpnet-base      & $83.2$ $(\uparrow +11.9)$ & $72.5$ $(\uparrow +1.2)$ & $84.9$ $(\uparrow8.5)$ & $72.7$ $(\downarrow -3.7)$\\
    distilroberta   & $75.8$ $(\uparrow +13.4)$ & $73.9$ $(\uparrow +11.5)$ & $71.5$ $(\uparrow +1.9)$ & $69.5$ $(\downarrow -0.1)$\\
    MiniLM-L12-v2   & $81.8$ $(\uparrow +19.09)$ & $72.4$ $(\uparrow +9.7)$ & $79.5$ $(\uparrow +8.6)$ & $70.2$ $(\downarrow -0.7)$\\
    MiniLM-L6       & $80.9$ $(\uparrow +15.1)$ & $72.7$ $(\uparrow +6.9)$ & $79.7$ $(\uparrow +9.2)$ & $69.3$ $(\downarrow - 1.2)$\\
    Para-albert     & $79.9$ $(\uparrow +10.3)$ & $68.8$ $(\downarrow -0.8)$ & $80.1$ $(\uparrow +13.6)$ & $74.6$ $(\uparrow +8.1)$\\
    Para-MiniLM-L3  & $81.6$ $(\uparrow +18.4)$ & $67.7$ $(\uparrow +4.5)$ & $76.8$ $(\uparrow +9.9)$ & $70.2$ $(\uparrow +3.3)$\\
    \bottomrule
  \end{tabular}
  \caption{Results for the experiment showing that models are sensitive towards specific words and cannot properly leverage the context, in the form of a definition. The figures in brackets indicate the difference in accuracy with respect to the experiments with prompts only of table~~\ref{tab:prompt_sensitivity_results}.}
  \label{tab:using_definition}
\end{table*}

\subsubsection{Inability to model negation}
\label{subsubsection_inability_to_model_negation}

Despite the acquired grammatical understanding reported by LLMs, they cannot model negation (e.g. ``not rock'')~\cite{negation_dataset_MLP_garcia_2023}. To validate if this holds for musical labels, we constructed a separate list of triplets for both ``Musical Genre'' and ``Musical Instruments''. For each valid triplet obtained, we extracted unique anchor-positive pairs and introduced a negative label as a negation of the anchor and positive labels. We are left with 3756 and 8284 negative triplets for Genres and Instruments respectively. These were then used alongside 4 negative prompts, listed in appendix~\ref{sec:negation-modelling}.

The performance is worse than random, as shown in table~\ref{tab:prompt_sensitivity_results}, which provides further evidence that LLMs cannot model negation in general and musical terminology. Different prompts lead to considerable differences in accuracy, with the worst performance reported being $\approx 23\%$. This might have potential implications in applications such as captioning, as datasets include negation.

\subsubsection{Sensitivity towards the presence of specific words}
Using artificially generated definitions of labels instead of generic prompts led to an increased zero-shot image classification accuracy~\cite{pratt2023does}. Drawing inspiration from this and leveraging single-sentence definitions provided by Audioset, we evaluate the performance when using the label-free definition and the combination of the label and definition simultaneously.

Excluding the label from the definition leads to a drop in every experiment, meaning that models might be sensitive to labels and not the semantics provided indirectly by the definition. On the other hand, the definition leads to an increment in accuracy in most cases, as shown in table~\ref{tab:using_definition}.

\section{Conclusions and future work}

In this paper, we quantified the musical knowledge of six Transformer-based models based on triplet accuracy with musical labels for genres and instruments. We identified three shortcomings: prompt sensitivity, difficulty modeling negation and sensitivity to specific words.

To overcome these shortcomings, we propose using augmentation during training and varying the prompt structures to avoid prompt sensitivity. This approach can utilize definitions to substitute labels with their definitions.
To address negation modeling, we suggest multi-task learning that includes tagging negative labels in a caption and maximizing the distance between negative and positive versions of the tags in contrastive losses. 

We recommend using lexical databases (e.g. WordNet), which offer more elaborate music concept relationships, instead of using a tree to obtain triplets. We highlight that further filtering needs to be done to form meaningful triplets and produce good-quality evaluation datasets. Lastly, despite reporting increments when definitions are used, further testing is required.

\bibliography{nlp4MusA}

\appendix

\section{Acknowledgments} The first author is a research student at the UKRI Centre for Doctoral Training in Artificial Intelligence and Music, supported jointly by UK Research and Innovation [grant number EP/S022694/1] and Queen Mary University of London.

\section{Language models used}
\label{appendix:language_models_used}

All the models used are pretrained and then fine-tuned for sentence similarity on several corpora of pairs. Paraphrase models share the same fine-tuning dataset and the same happens for the remaining 4, with an additional 50 million sentence pairs for all-distilroberta-v1. More information can be found in the respective papers, Sentence Transformer~\footnote{https://sbert.net/} package documentation and Hugging Face websites~\footnote{https://huggingface.co/}.

MPNet unifies the Masked Language Modeling (MLM) and Permuted Language Modeling pretasks, used by BERT~\cite{devlin-etal-2019-bert} and XLNet~\cite{yang2019xlNet} respectively, to train a Transformer backbone. The tokens of the input are permuted, a set of them is masked and the objective is to predict the masked section, while the positional information of the full sentence is also known.

DistilBERT~\cite{sanh2020distilbertdistilledversionbert} is a 40\% smaller BERT model that is trained on the same regime as BERT but with an additional loss term. The distillation loss~\cite{Hinton2015DistillingTK} is:

\begin{equation}
    L_{ce} = \sum_{i} t_{i} * log(s_{i})
\end{equation}

where $t_{i}, s_{i}$ is the probability for the predicted tokens of the teacher (BERT) and student (DistilBERT) models respectively. This is used to let the student approximate the target probability distribution of the teacher and therefore, learn from the teacher model.

RoBERTa~\cite{Liu2019RoBERTaAR} is a model based on BERT with removing next-sentence prediction pretraining, increasing the mini-batch size and altering key hyperparamaters. The analysis of the last are out of the scope for this paper. DistilRoBERTa uses RoBERTa and the distillation process described for DistilBERT.

Instead of approximating the target probability distribution, MiniLM~\cite{minilm_wang_2020} proposed to ``mimic'' the last self-attention module between the student and teacher models. In addition to approximating the attention distribution, this system approximates the relations between the scaled dot-products of queries, keys and value embeddings. Therefore, it also models the second-degree associations between the self-attention embeddings, as well as their distribution.

Finally, ALBERT~\cite{Lan2019ALBERTAL} utilizes parameter reduction techniques, as well as swapping the Next Sentence Prediction to Sentence Ordering Prediction. Firstly, Factorized Embedding Parametrization is used to decompose the vocabulary embedding matrix into two small matrices. As a result, the size of the hidden layers is decoupled from the size of the token embeddings. Secondly, Cross-Layer Parameter Sharing relaxes the dependency between memory demands and model depth. Lastly, Sentence Ordering Prediction is focused on predicting the sequence of two sentence segments, while Next Sentence Prediction is used to predict if the pair of sentences is from the same document or not.

\section{Prompts used}
\subsection{Prompt sensitivity}
\label{appendix:prompt_sensitivity}
The prompts used for evaluating the sensitivity towards different musically informed prompts of Transformer-based models are:

\begin{figure}
    \centering
    \includegraphics[width=1\linewidth]{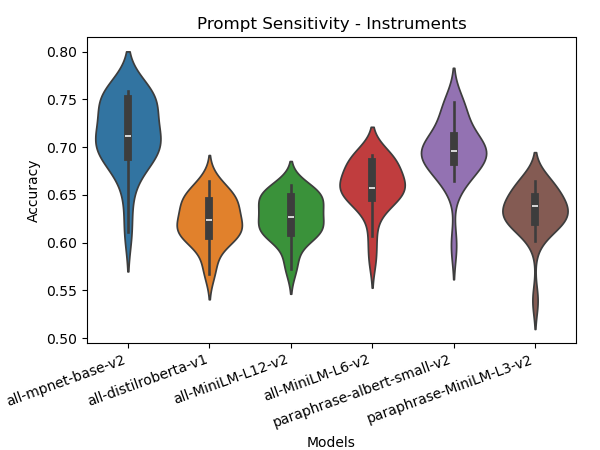}
    \caption{Prompt sensitivity of 6 Transformer-based models with respect to musical instrument terminology.}
    \label{fig:prompt_sensitivity_instruments}
\end{figure}

\begin{enumerate}
    \setlength\itemsep{0em}
    \item ``\textit{The sound of <label>}''
    \item ``\textit{Music made with <label>}''
    \item ``\textit{A <label> track}''
    \item ``\textit{This is a recording of <label>}''
    \item ``\textit{A song with <label>}''
    \item ``\textit{A track with <label> recorded}''
    \item ``\textit{A music project with <label>}''
    \item ``\textit{Music made from <label>}''
    \item ``\textit{Music of <label>}''
    \item ``\textit{A music recording of <label>}''
    \item ``\textit{This song is made from <label>}''
    \item ``\textit{The song has <label>}''
    \item ``\textit{Music song with <label>}''
    \item ``\textit{Music song with <label> recorded}''
    \item ``\textit{Musical sounds from <label>}''
    \item ``\textit{This song sounds like <label>}''
    \item ``\textit{This music sounds like <label>}''
    \item ``\textit{Song with <label> recorded}''
    \item ``\textit{A <label> music track}''
    \item ``\textit{Sound of <label>}''
\end{enumerate}

\begin{figure}
    \centering
    \includegraphics[width=1\linewidth]{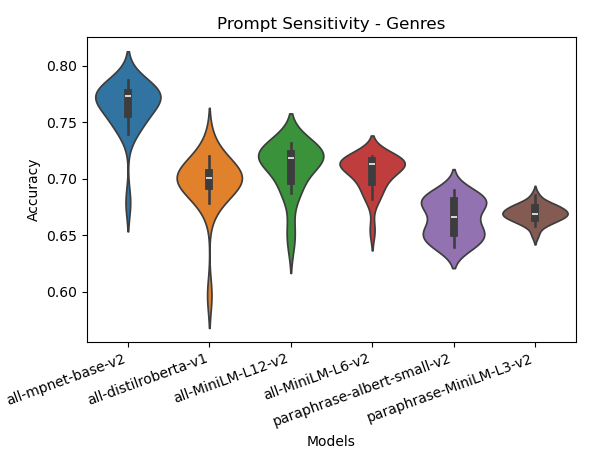}
    \caption{Prompt sensitivity of 6 Transformer-based models with respect to musical genre terminology.}
    \label{fig:prompt_sensitivity_genres}
\end{figure}

\subsection{Negation modeling}\label{sec:negation-modelling}
The four prompts used to evaluate the inability to model negation:

\begin{enumerate}
    \setlength\itemsep{0em}
    \item ``\textit{No <label>}''
    \item ``\textit{Not the sound of <label>}''
    \item ``\textit{Doesn't sound like <label>}''
    \item ``\textit{Not music from <label>}``
\end{enumerate}

\begin{table}[ht!]
    \begin{subtable}{\linewidth}
    \centering
    \begin{tabular}{lcccc}
      \toprule
      \textbf{} & \multicolumn{4}{c}{\textbf{Instrument Prompts}}\\
      \textbf{Models} & \textbf{\#1} & \textbf{\#2} & \textbf{\#3} & \textbf{\#4}\\
      \midrule
      mpnet-base      & $45.4$ & $35.9$ & $44.0$ & $39.5$\\
      distilroberta   & $41.4$ & $31.6$ & $39.0$ & $36.7$\\
      MiniLM-L12-v2   & $44.2$ & $30.8$ & $33.6$ & $26.8$\\
      MiniLM-L6       & $46.9$ & $33.2$ & $37.5$ & $32.0$\\
      Para-albert     & $42.7$ & $28.6$ & $28.5$ & $34.0$\\
      Para-MiniLM-L3  & $40.4$ & $23.6$ & $26.8$ & $25.1$\\
      \bottomrule
    \end{tabular}
    \caption{Instruments}
    \label{tab:negation_modeling_instruments}
    \end{subtable}
    
  \vfill

    \begin{subtable}{\linewidth}
    \centering
    \begin{tabular}{lcccc}
      \toprule
      \textbf{} & \multicolumn{4}{c}{\textbf{Genre Prompts}}\\
      \textbf{Models} & \textbf{\#1} & \textbf{\#2} & \textbf{\#3} & \textbf{\#4}\\
      \midrule
      mpnet-base      & $49.0$ & $39.5$ & $44.0$ & $40.0$\\
      distilroberta   & $45.6$ & $37.8$ & $45.2$ & $40.1$\\
      MiniLM-L12-v2   & $47.2$ & $35.6$ & $38.7$ & $27.9$\\
      MiniLM-L6       & $49.6$ & $40.7$ & $42.1$ & $33.3$\\
      Para-albert     & $44.8$ & $32.2$ & $29.8$ & $35.7$\\
      Para-MiniLM-L3  & $42.4$ & $32.5$ & $33.0$ & $29.1$\\
      \bottomrule
    \end{tabular}
    \caption{Genres}
    \label{tab:negation_modeling_genres}
    \end{subtable}

  \caption{Presentation of results for experiment~\ref{subsubsection_inability_to_model_negation}. No model performed on par with the random baseline.}
  \label{tab:negation_modeling}
\end{table}

\subsection{Examples of removed triplets}
As stated in~\ref{subsection_triplet_based_musical_knowledge_quantification}, there were some triplets of ambiguous quality. We argue that removing these is far more important than building a very big evaluation dataset.

For reference, we present 10 triplets of different ambiguousness levels for each category in table~\ref{tab:ambiguous_triplets}.

\begin{table*}[]
    \centering
    \begin{tabular}{c c c}
        \toprule
        \multicolumn{3}{c}{\textbf{Instruments}}\\
        \midrule
         Anchor & Positive & Negative\\
         \midrule
         Musical instrument & Plucked string instrument & Mandolin\\
        Cowbell & Accordion & Flute\\
        Guitar & French horn & Timpani\\
        Electric guitar & Hammond organ & Rhodes piano\\
        Bass guitar & Brass Instrument & Alto saxophone\\
        Tapping (guitar technique) & French horn & Electric piano\\
        Sitar & Cymbal & Rimshot\\
        Keyboard (musical) & Cowbell & Acoustic guitar\\
        Piano & Didgeridoo & Cello\\
        Organ & Trombone & Timpani\\

        \bottomrule
        \toprule
        \multicolumn{3}{c}{\textbf{Genres}}\\
        \midrule
         Anchor & Positive & Negative\\
        \midrule
        Music genre & Rhythm and blues & Swing music\\
        Pop music & Jazz & Swing music\\
        Hip hop music & Classical music & Drum and bass\\
        Rock music & Independent music & Grime music\\
        Heavy metal & Electronic music & Oldschool jungle\\
        Progressive rock & Chant & Oldschool jungle\\
        Reggae & Music of Asia & Cumbia\\
        Jazz & New-age music & Heavy metal\\
        Kuduro & Music for children & Grunge\\
        Funk carioca & Christian music & Electronica\\
        \bottomrule
    \end{tabular}
    \caption{Table with examples of removed triplets. The filtering criterion is based on the ambiguity or relative difficulty in determining whether the anchor is more similar to the positive or negative label.}
    \label{tab:ambiguous_triplets}
\end{table*}

\section{Detailed experiment results}
\label{appendix:detailed_experiment_results}
\subsection{Prompt sensitivity}
Generally, prompt sensitivity is evident in every model. The biggest and best model, all-mpnet-base-v2, has the largest and one of the largest variances for instruments (figure~\ref{fig:prompt_sensitivity_instruments}) and genres respectively (figure~\ref{fig:prompt_sensitivity_genres}). 

Paraphrase-MiniLM-L3-v2 had the smallest variance for genres, at the expense of a lower accuracy. This might be due to the different distillation process chosen. If an application demands robustness towards prompt sensitivity, that would be the best choice.

Apart from all-mpnet-base-v2, every model had approximately the same variance when the outliers were discarded, as can be seen in figure~\ref{fig:prompt_sensitivity_instruments}.

\subsection{Negation modeling}
By far the worst deficiency found is the inability of Transformer-based models to model negation. These failed to surpass random choice in every experiment, while altering the prompt led to a significant decrease in accuracy, up to $\approx20\%$. This is presented in table~\ref{tab:negation_modeling_instruments}.

This result can have large implications on developing or evaluating captioning systems, as datasets~\cite{Agostinelli2023MusicLMGM, manco2023thesong} contain negation and following these results, can lead to erroneous inference. Also, joint audio-text models, also known as two-tower systems, can be negatively impacted. Further testing is required in the future.

\end{document}